# Hand Gesture Detection and Conversion to Speech and Text

K. Manikandan, Ayush Patidar*, Pallav Walia*, Aneek Barman Roy*

Department of Information Technology, SRM Institute of Science and Technology,
manikandan.k@ktr.srmuniv.ac.in, {ayushpatidar_tulsiram, pallav_walia, aneek_suman}@srmuniv.edu.in

## ABSTRACT

The hand gestures are one of the typical methods used in sign language. It is very difficult for the hearing-impaired people to communicate with the world. This project presents a solution that will not only automatically recognize the hand gestures but will also convert it into speech and text output so that impaired person can easily communicate with normal people. A camera attached to computer will capture images of hand and the contour feature extraction is used to recognize the hand gestures of the person. Based on the recognized gestures, the recorded soundtrack will be played.

## KEYWORDS

Speech, Hand Gesture, Hull, Image Processing

## I. INTRODUCTION

As computer innovation keeps on developing, the requirement for characteristic correspondence amongst people and machines additionally increments. In spite of the fact that our cell phones influence utilization of the touch to screen innovation, it is not sufficiently shabby to be actualized in work area frameworks. In spite of the fact that the mouse is exceptionally valuable for gadget control, it could be badly arranged to use for physically disabled individuals and individuals who are not familiar with utilizing the mouse for connection. The strategy proposed in this paper makes utilization of a webcam through which hand gestures gave by the user are captured and identified accordingly. Hand gestures have boundless applications. In this investigation, we apply it to a system to make a straightforward easy to understand interaction interface.

The research paper sheds light on the recognition of letters from the hand language which is taken as per the American Sign Language. The detection is done using the various techniques of Contour Analysis and Feature Extraction.

The research paper sheds light on the recognition of letters from the hand language which is taken as per the American Sign Language. The detection is done using the various techniques of Contour Analysis and Feature Extraction.

The paper invokes the use of various computer vision techniques and algorithms which are involved in the determination of hand gestures.

## II. LITERATURE REVIEW

In the past, many techniques have been used to convert the hand gesture to text. However, they were limited in terms of their functionalities. Many techniques required gloves with sensors which not only made the application more complex but also expensive. In the other version, the system was limited to a particular background without any noise or disturbance. There were some projects which were heavily dependent on heavy GPUs making it difficult for common man to use the system. Additionally, there were some systems for detections which required the object to be of a particular skin colour. Although, there have been various techniques for converting the hand gesture to text but a very few focus on converting the gesture to both text and speech with that too with limited properties.

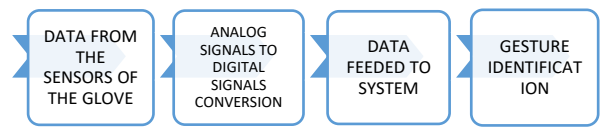

## III. SOLVING THE PROBLEM

The objective of this project is to provide a tool to aid the speech impaired by retrieving the features of the hand. This proposal has been developed using OpenCV with properly designed and user-friendly interface.

## IV. SYSTEM ARCHITECTURE

In the present world, managing diversity is under threat, especially when impaired people are involved.

The present machine usage focuses on its efficiency and its sophisticated design. The recent modules involve usage of equipment viz. gloves, pointers to help in hand movement in applications involving communication.

This algorithm can be used by the people who are deaf-mute. It provides a medium for these people to interact with the outside world. It can be used to convert hand gestures both into speech and text. Not only it can be used for the visually impaired, but also by a common man to convert his actions into text or speech based on his requirement.

The proposed solution falls between visually based interfaces. The image of the hand is taken through the camera and is converted into a binary image. The binary image is basically monochromatic and can be processed. Hence, the OpenCV library is optimized and used for such actions.

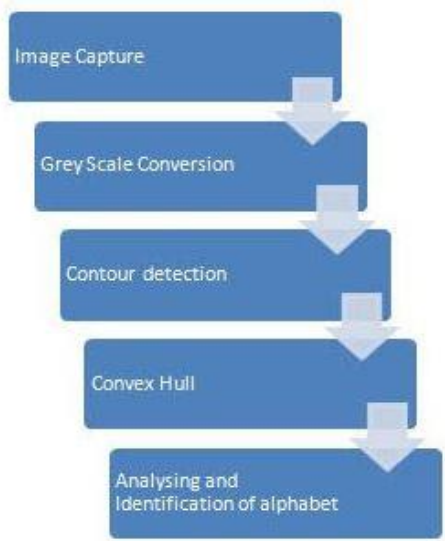

The OpenCV consists of a function which determines the edge of the object and is defined as the Contour of the object. For instance, the contour of a tennis court is a rectangle. Various solutions for extracting edges in a binary image were suggested and one of the primitive algorithms is considered as a foundation. There are certain algorithms in OpenCV library that provide various techniques for finding the features of a contour. The extraction of points of the contour speed up the overall process of further contour processes. The said set of points integrate to form into an n-dimensional polygon which is known as the hull.

The shape of the hull can be concave or convex in nature. Sometimes, it is not possible to draw a line inside the polygon and it would intersect its border. Such a hull is called convex hull. If not, then the hull is not convex in nature and contains convexity defects.

The hand does have huge convexity defects between the fingers. Hence such area properties will help in the design and analysis of the algorithm.

V. SOLUTION AND RESULTS

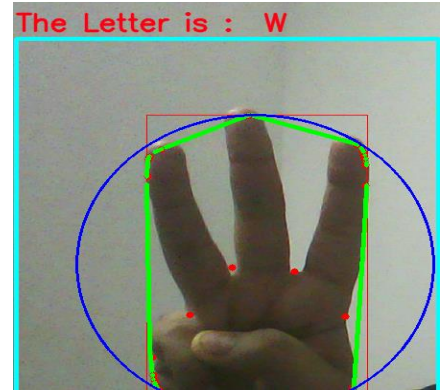

Figure1. Hand Gesture of letter W

*A. Creating the Binary Image*

The first step in hand gesture recognition is to create a threshold of the image. It is necessary that we separate the hand gesture region with the background so as to make the process of recognition a bit simpler. This is based on the different pixel intensity of the hand and the background resulting in the separation of the two. After separation, we assign the important pixels (hand region) with a value to identify them. Like the value of 255(white), 0(black), or any other value accordingly.

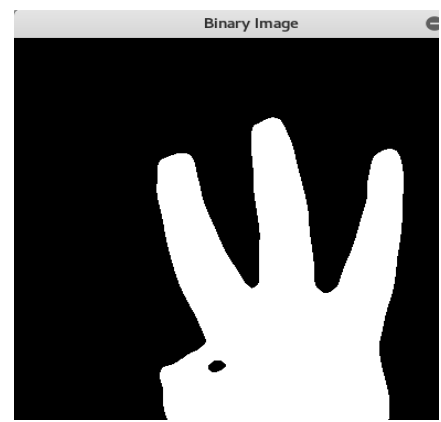

Figure 2. Binary image of Letter "W"

*B. Finding the outline of the image using contour.*

The shape of the hand is identified using inbuilt OpenCV functions through contours detection technique. The function will return the set of coordinates of contours which will help in drawing the complete shape. Finding the outline also helps in determining the various contour properties that are necessary for the identification of letters.

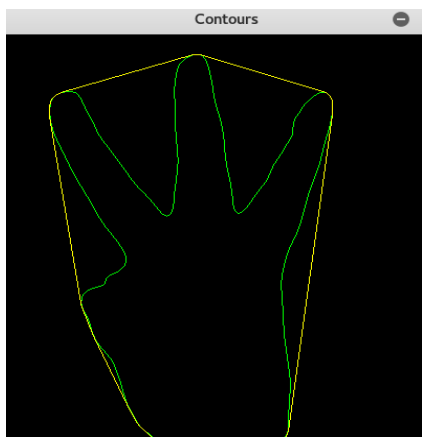

Figure 3. Hand contour with the identification of convex hull.

### C. Calculating convex Hull and Defects

Convexity defect is a cavity in an object that means an area that does not belong to the object but located inside of its outer boundary (convex hull). From the contour analysis, we use to get the data which later is manipulated to find the Number of Convexity Defects. Based on the numbers we can identify the number of fingers which later is used to identify the letter.

The number of contours is calculated as follows

We register a triangle. Give the sides a chance to be a, b and c. This triangle is shaped by the beginning stage of the form, the consummation purpose of the form and the most remote purpose of the shape. (a, b, c separately). "a" is computed as follows

a = math.sqrt((end[0] - start[0])**2 + (end[1] - start[1])**2) [7]
b = math.sqrt((far[0] - start[0])**2 + (far[1] - start[1])**2)
c = math.sqrt((end[0] - far[0])**2 + (end[1] - far[1])**2)

Now, using the Cosine rule ,

$$\cos(A) = (b^2 + c^2 - a^2)/2bc$$

$$\cos(B) = (a^2 + c^2 - b^2)/2ac$$

$$\cos(C) = (a^2 + b^2 - c^2)/2ab$$

The angle A is calculated.

If the angle A is less than or equal to 90 degrees, it means that there is a convexity defect. Once there is a convexity defect recognized, a variable by the name, cnt, increments by one. So, by this algorithm, we can efficiently identify how many convexity defects there are.

### D. Identification of Letters

Alphabet A**:** Alphabet A can be identified by computing the difference between the area of a circle and the area of the contour. The circle is obtained by bounding the contours. In case of alphabet A only there is a very little difference between two areas. Hence this method is proved to be efficient to identify A.

Alphabet B: For the alphabet B, We only need to compute the contour area as it has the largest area among other alphabets.

ALPHABETS V, C, L, and Y: The letters come into picture when the identification of letter A fails. If the number of convexity defects is equal to 1 then the parameter Angle is calculated. The angle is obtained by an OpenCV inbuilt function that calculates the overall figure's orientation. Based on the value of the angle.

ALPHABETS F and W: The only two alphabets (F and W) in American Sign Language which have two convexity defects. If there are two convexity defects then we compare the angle. From this the F and W can be identified easily.

ALPHABETS D, J, H, I, U: These alphabets requires the combination of parameters which include Solidity, Aspect Ratio and Angles. The above parameters are found to be reliable after the intensive testing.

| ALPHABETS | ANGLE RANGE |
|---|---|
| V | <10 |
| C | 40-66 |
| L | 20-35 |
| F | >100 |
| D | <20 |
| I | 169-180 |
| H | 30-100 |
| J | <168 |

Table1. The corresponding angles of various letters

E. *Examining the features of a contour.*

Various Contour properties are computed to identify the letter made. They are listed as follows

1) The Solidity of a contour. It is defined as the ratio of contour area to the area of the convex hull.

2) The angle within a contour. It is the region between the rays of the alphabet and is the property used to determine the letter.

3) The diameter of Contour Area-It the line passing through the centre of the letter and can be used to determine the type of letter with a circular shape.

4) The arc length of Contour. It is the Perimeter of Contour is defined as the overall perimeter of the letter.

5) A Number of Defects- It is basically the cavity which is present in an object. It is in the insignificant area in an object.

6) The Moments within a contour. Moments are useful for describing the vectors and spatial distribution present in an image

7) The Bounding Rectangle Area covering the contour. The bounding rectangle area is the rectangular region enclosing the letter or the alphabet, required to be identified.

8) The overall area of Contour- The area of contour is the area formed by the letter or the alphabet and is very useful in determining the type of letter.

F. *Speech Conversion Of The Identified Letter*

Once we identify the letter, the test was to play a sound file which played "The Letter will be (Letter perceived). The audio file wasn't getting completed once played because the frame the stream of the video constantly changes. To overcome this issue, a bash script was written to handle the audio part. After the Letter is identified, a defined function is executed and run. For instance, the alphabet U is detected and a directory is made and navigated into that directory. Then the .txt file is made containing the alphabet U. If it already exists, it is overwritten.

A Shell Script is written which goes into the directory to find the latest modified file. After finding the file, the corresponding audio file is played using the script. This script is completely independent of the python program. The individual sound files are stored in the system. Once the letter is identified, the corresponding the file is played after running the script. This ensures that the sound is played continuously and without any disturbance.

VI. TRAIL EFFICIENCY

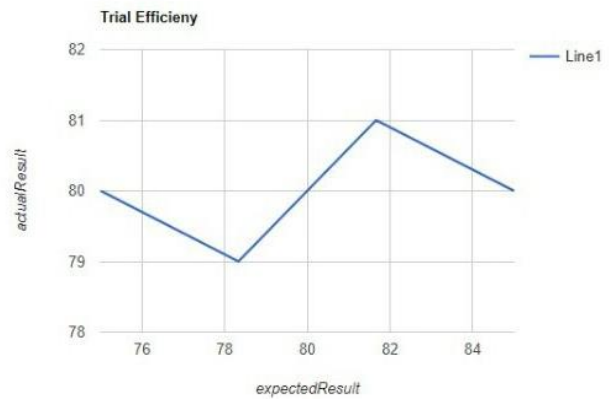

Efficiency is a key measure of any solution model or hypothesis. The efficiency can be measured in various ways. However, this solution requires a hit and trial method because of the interaction between the system and the object (hand in this case).

The two scale of measures is expected output and the actual output. The expected output is defined to be in the range of 75% to 85% as optimum usage of algorithms is present. The actual output is calculated as a measure of the above-mentioned hit and trial method.

Let actual output be defined as "$a_o$".

$$a_o=(\text{time taken in seconds})/7 \text{ seconds} *100$$

We have used 7 seconds as a standardized value to deduce able comparisons. The persistence and the ability of the system to function well will hold true for such a value.

Based on the above test hypothesis and models, four rounds of tests are done and the line graph has been plotted by putting the actual output efficiency vs the resultant output efficiency.

G. . APPLICATION DOMAIN

There are two programs. One for the speech impaired and one for the visually impaired.

Speech Impaired:

The people who cannot speak can simply communicate with the outside world with the help of sign language. Using the hand gestures, a person can convey the message. The desired message is converted to both text and speech.

Hearing Impaired:

The deaf people can make use of this system to communicate. The person can simply use the hand movement to convert it into a text message which is further displayed on the screen.

## VII. FUTURE SCOPE

The application can be integrated with other mobile and IoT devices to improve user interaction and make the system more robust. The accuracy of the program can be further improvised by using neural networks.

An alternate stress could be put on the use of the application in the fields of medicines, military, governance etc. A genuine blend of various technologies in mentioned fields could make way for power tools and applications which will serve the community around the world.

Finally, the use can be further designed to make more accessible to the consumers. The whole point of making the solution as a commercially viable product for the users is to help the impaired community around the world.

## VI. CONCLUSION

The practical adaption of the interface solution for visually impaired and blind people is limited by simplicity and usability in practical scenarios. As an easy and practical way to achieve human-computer- interaction, in this solution hand gesture to speech and text conversion has been used to facilitate the reduction of hardware components.

On the whole, the solution aims to provide aid to those in need thus ensuring social relevance. The people can easily communicate with each other. The user-friendly nature of the system ensure that people can use it without any difficulty and complexity. The application is cost efficient and eliminates the usage of expensive technology.